%% file: main.tex
\documentclass[accepted]{uai2021} 

\usepackage[american]{babel}

\usepackage{natbib} 
\usepackage{mathtools} 
\usepackage{booktabs} 
\usepackage{tikz} 

\include{macros}

\title{Sketching Curvature for Efficient Out-of-Distribution \\ Detection for Deep Neural Networks}

\author[1]{\href{mailto:Apoorva Sharma <apoorva@stanford.edu>?Subject=SCOD}{Apoorva Sharma}{}}
\author[1,2]{Navid Azizan}
\author[1]{Marco Pavone}
\affil[1]{%
    Stanford University
}
 \affil[2]{%
    Massachusetts Institute of Technology
}

\begin{document}
\maketitle

\begin{abstract}
\input{0_abstract}
\end{abstract}

\section{Introduction}
\label{intro}
\input{1_intro}

\section{Problem Statement}
\label{prob_statment}
\input{2_problem_statement}

\section{Background: Curvature and Laplace Approximation}
\label{prelim}
\input{3_preliminaries}

\section{Proposed Method: \algNameLong{}}
\label{approach}
\input{4_approach}

\section{Related Work}
\label{relatedwork}
\input{5_related_work}

\section{Experimental Results}
\label{experiments}
\input{6_experiments}

\section{Conclusion}
\label{conclusion}
\input{7_conclusions}



\begin{acknowledgements} 
    A.S. and M.P. were supported in part by DARPA under the Assured Autonomy program and by NASA under the University Leadership Initiative program. N.A. was supported by MIT. The authors wish to thank Robin Brown and Edward Schmerling for helpful input and discussions during the development of these ideas. 
\end{acknowledgements}
\bibliography{biblio}

\clearpage
\appendix
\input{A_appendix}

\end{document}

%% file: macros.tex
\usepackage{bm}
\usepackage{amsmath}
\usepackage{amssymb}
\usepackage{nicefrac}
\usepackage{algorithm}
\usepackage{algpseudocode}
\usepackage{multirow}

\renewcommand{\vec}[1]{\bm{#1}}

\newcommand{\T}{\mathsf{T}} 

\newcommand{\x}{\bm{x}}
\newcommand{\inputspace}{\mathcal{X}}
\newcommand{\y}{\bm{y}}
\newcommand{\outputspace}{\mathcal{Y}}

\newcommand{\test}{{(t)}}

\newcommand{\ndata}{M}
\newcommand{\nparam}{N}
\newcommand{\param}{\bm{w}}
\newcommand{\distparam}{\bm{\theta}}
\newcommand{\ndistparam}{d}
\newcommand{\dataset}{\mathcal{D}}
\newcommand{\R}{\mathbb{R}}
\newcommand{\subspacedim}{k}
\newcommand{\subspacebasis}{U_\mathrm{top}}

\newcommand{\FIM}{F}

\newcommand{\Jac}{\bm{J}}
\newcommand{\rank}{\mathsf{rank}}
\newcommand{\diag}{\mathsf{diag}}
\DeclareMathOperator{\trace}{Tr}

\newcommand{\KL}{D_\mathsf{KL}}
\newcommand{\DeltaKL}{\delta_\mathsf{KL}}
\newcommand{\E}{\mathbb{E}}
\newcommand{\oodmetric}{\mathrm{Unc}}

\newcommand{\sketchop}{\mathsf{S}}

\newcommand{\N}{\mathcal{N}}

\newcommand{\algName}{SCOD}
\newcommand{\algNameLong}{Sketching Curvature for OoD Detection}

%% file: 0_abstract.tex
In order to safely deploy Deep Neural Networks (DNNs) within the perception pipelines of real-time decision making systems, there is a need for safeguards that can detect out-of-training-distribution (OoD) inputs both efficiently and accurately.
Building on recent work leveraging the local curvature of DNNs to reason about epistemic uncertainty,
we propose \emph{\algNameLong{} (\algName{})}, an architecture-agnostic framework for equipping any trained DNN with a task-relevant epistemic uncertainty estimate.
Offline, given a trained model and its training data, \algName{} employs tools from matrix sketching to tractably compute a low-rank approximation of the Fisher information matrix,
which characterizes which directions in the weight space are most influential on the predictions over the training data.
Online, we estimate uncertainty by measuring how much perturbations orthogonal to these directions can alter predictions at a new test input.
We apply \algName{} to pre-trained networks of varying architectures on several tasks, ranging from regression to classification. We demonstrate that \algName{} achieves comparable or better OoD detection performance with lower computational burden relative to existing baselines.

%% file: 1_intro.tex
Deep Neural Networks (DNNs) have enabled breakthroughs in extracting actionable information from high-dimensional input streams, such as videos or images. 
However, a key limitation of these black-box models is that their performance can be erratic when queried with inputs that are significantly different from those seen during training.
To alleviate this, there has been a growing field of literature aimed at equipping pre-trained DNN models with a measure of their uncertainty.
These approaches aim to characterize what a given DNN model has learned from the dataset it was trained on, so as to detect at test time whether a new input is inconsistent.

One appealing direction to this end has leveraged the curvature of a pre-trained DNN about its optimized weights \citep{madras_detecting_2019,Ritter2018ASL}.
The second-order analysis of the local sensitivity of the network to its weights can offer a post-hoc approximation of the intractable Bayesian posterior on the network weights. Known as the Laplace approximation, this is an attractive approach in its promise of adding a principled measure of epistemic uncertainty to any pre-trained network. However, computing the required curvature matrix is quadratic in the number of weights of the network, and is still intractable for today's DNN models with millions of weights. Thus, the literature has focused on methods to approximate this curvature to yield scalable approaches, yet these approaches have typically relied on imposing sparsity structures on the curvature matrix, e.g., by ignoring cross terms between layers of the network. Furthermore, these approaches tend to focus on estimating the \textit{posterior predictive} distribution by marginalizing over the approximate posterior over the weights, which often requires the computationally intensive process of sampling weights and estimating the posterior predictive through Monte-Carlo integration.

\begin{figure*}
    \centering
    \includegraphics[width=\textwidth]{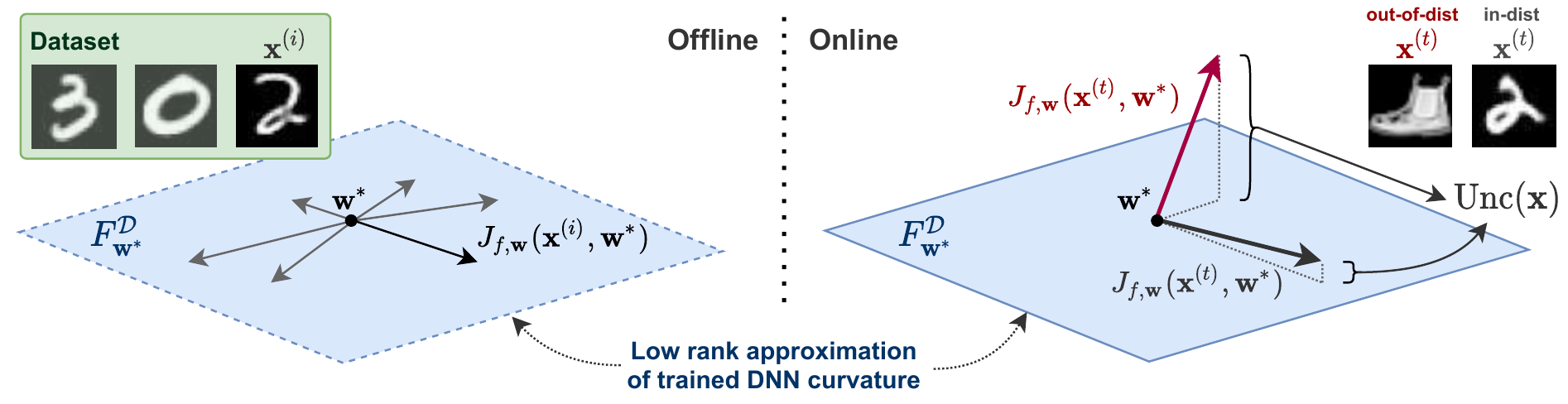}
    \caption{Offline, \algName{} uses gradients $\Jac_{f,\param}$ of the DNN on training data to build a low-rank estimate of the network curvature, as defined by the Fisher information matrix, employing tools from matrix sketching to make this estimation tractable and scalable. Online, \algName{} produces a metric of atypicality $\oodmetric$ using gradient information at a test datapoint $\x^\test$, and comparing this to the low-rank approximation of the Fisher. Derived from an approximation of the Bayesian posterior over the weights, this metric can be viewed, for scalar regression DNNs, as measuring how orthogonal a new test point gradient is from the gradients seen on training inputs.}
    \label{fig:overview_fig}
\end{figure*}

\paragraph{Contributions.}
In this work, we build from the framework of the Laplace approximation to propose a novel, architecture-agnostic method for equipping any trained DNN with estimates of task-relevant input atypicality.
Rather than incorporate epistemic uncertainty into the probabilistic prediction of the network, we propose augmenting the network output with a scalar uncertainty measure that quantifies the degree of atypicality for a given input. 
Our specific contributions are as follows:
    (1) We leverage information geometry to translate curvature-informed weight uncertainty in DNNs to a task-relevant measure of input atypicality;
    (2) we show how this measure can be computed efficiently by leveraging backpropagation and a characterization of the top eigenspace of the Fisher matrix;
    (3) we develop a technique based on matrix sketching to tractably approximate this top eigenspace, even for large models and large datasets;
    (4) we empirically demonstrate that this atypicality measure matches or exceeds the OoD detection performance relative to baselines on a diverse set of problems and architectures from regression to classification.

%% file: 2_problem_statement.tex
We take a probabilistic interpretation, and define a DNN model as a mapping from inputs $\x \in \inputspace$ to probability distributions over targets $\y \in \outputspace$. Typically, this is broken down into the composition of a neural network architecture $f$ mapping inputs $\x$ and weights $\param \in \R^\nparam$ to parameters $\distparam \in \R^\ndistparam$, and and a distributional family $\mathcal{P}$ mapping these parameters to a distribution over the targets:
\begin{align*}
    \distparam = f(\x, \param), \quad\quad p(\y) = \mathcal{P}(\distparam).
\end{align*}
We use $p_{\param}(\y \mid \x) := \mathcal{P} \circ f$ as a convenient shorthand for the conditional distribution that the DNN model defines.
This model is trained on a dataset of examples $\dataset = \{ \x_i, \y_i \}_{i=1}^{\ndata}$, where we assume the inputs are drawn i.i.d. from some training distribution $p_\mathrm{train}(x,y)$, to minimize the Kullback-Leibler (KL) divergence from the empirical distribution in the dataset to $p_{\param}(\y \mid \x)$
\footnote{This general formulation covers almost all applications of DNNs; for example, choosing $\mathcal{P}$ to be a categorical distribution parameterized by logits $\distparam$ recovers the standard softmax training objective for classification, while choosing $\mathcal{P}$ to be unit variance Gaussian with mean $\distparam$ recovers the typical mean-squared error objective for regression.}.

Our goal is to take a trained DNN, and equip it with an OoD monitor which can detect if a given input is far from the data seen at train time, and thus likely to result in incorrect and overconfident predictions. Specifically, we assume we are given the functional forms $f$ and $\mathcal{P}$, the trained weights $\param^*$, as well as a set of training data $\dataset$, and wish to construct an uncertainty measure $\oodmetric(\cdot): \inputspace \rightarrow \R$. Intuitively, we wish for this measure to be low when queried on inputs drawn from $p_\mathrm{train}(x)$, but high for inputs far from this data manifold. This can be thought of as providing a measure of \textit{epistemic uncertainty} due to an input not being covered by the training set, or a measure of \textit{novelty} with respect to the training data, useful for detecting \textit{anomalies} or \textit{out-of-distribution} inputs. 

We wish to design a monitor that can provide a real-time uncertainty signal alongside the DNN's predictions. Thus, we desire a function $\oodmetric(\cdot)$ that is both computationally efficient (to run at test time) as well as informative, providing a meaningful anomaly signal that can effectively separate held-out validation inputs that are known to be in-distribution, from inputs which we know to be out-of-distribution and outside the DNN's domain of competency.

%% file: 3_preliminaries.tex
Key to our approach is the curvature of DNN models: a second-order characterization of how perturbations to the weights of a DNN influence its probabilistic output. In this section, we review tools to characterize DNN curvature, and discuss the connections between curvature and epistemic uncertainty estimates through the Laplace approximation.

\subsection{Fisher Information}
A crucial aspect of our approach relies on understanding how the parameters of the model influence its output distribution. To do so, we leverage tools from information geometry. We can view the family of distributions defined by the model $\mathcal{P}(\distparam)$ as defining a statistical manifold with coordinates $\distparam$. The Riemannian metric for this manifold is the Fisher information matrix
\begin{align}
    \FIM_{\distparam}(\distparam) &= \E_{\y} \left[ ( \nabla_{\distparam} \log p_{\distparam}(\y) )(\nabla_{\distparam} \log p_{\distparam}(\y))^\T \right] \label{eq:fisher_def} \\
     &= \E_{\y} \left[ - \frac{\partial^2}{\partial \distparam^2} \log p_{\distparam}(\y) \right], \quad\text{(see note)\footnotemark}
\end{align} \footnotetext{This equality holds under mild regularity conditions on $\mathcal{P}(\distparam)$}
where $p_{\distparam}(\cdot) = \mathcal{P}(\distparam)$, the pdf of the probability distribution on $\outputspace$ defined by parameters $\distparam$.
The Fisher information matrix (henceforth referred to as the Fisher) represents the second-order approximation of the local KL divergence, describing how the output distribution of a model changes with small perturbations to the distribution parameters $\distparam$:
\begin{align}
    \KL(\mathcal{P}(\distparam) || \mathcal{P}(\distparam + d\distparam)) &\approx d\distparam^\T \FIM_{\distparam}(\distparam) d\distparam + O(d\distparam^3). \label{eq:fisher-kl}
\end{align}
The subscript on $\FIM_{\distparam}$ serves to make explicit the Fisher's dependence on the model's parameterization. 

For many common parametric distributions, the Fisher can be computed analytically.
For example, for the family of Gaussian distributions with fixed covariance $\Sigma$, parameterized by the mean vector $\distparam = \bm{\mu}$, the Fisher information is simply the constant $\FIM_{\distparam}(\distparam) = \Sigma^{-1}$. For a categorical distribution parameterized such that $\distparam_i$ represents the probability assigned to class $i$, $\FIM(\distparam) = \left(\diag(\distparam)\right)^{-1}$. 
In cases where this analytic computation is not possible or difficult, one can compute a Monte-Carlo approximation of the Fisher by sampling $\y \sim p_{\distparam}(\cdot)$ to estimate the expectation in \eqref{eq:fisher_def}.

\subsection{Fisher for Deep Neural Networks}
For DNNs, we can also consider the Fisher in terms of the network weights $\param$ using a change of variables. Since $\distparam = f(x, \param)$, we have
\begin{align}
    \FIM_{\param}(\x, \param) &= \Jac_{f,\param}^\T \FIM_{\distparam}(f(\x, \param)) \Jac_{f,\param}, \label{eq:reparameterization}
\end{align}
where $\Jac_{f,\param}$ is the Jacobian matrix with $[\Jac_{f,\param}]_{ij} = \frac{\partial f_i}{\partial w_j}$, evaluated at $(\x, \param)$. Note that $\FIM_{\param}$ is a function of both $\param$ and the input $\x$. We will henceforth use the shorthand $\FIM^{\test}_{\param^*} := \FIM_{\param}(\x^\test, \param^*)$ to denote this weight-space Fisher evaluated for a particular input $\x^\test$ and the trained weights $\param^*$. 

From \eqref{eq:fisher-kl}, we see that the Fisher defines a second-order approximation of how perturbations in the weight space influence the DNN's probabilistic predictions:
\begin{align*}
    \DeltaKL(\x^\test) &:= \KL\left(p_{\param^*}(\cdot \mid \x^\test) || p_{\param^* + d\param}(\cdot \mid \x^\test) \right) \\
    &\approx d\param^\T \FIM^\test_{\param^*} d\param.
\end{align*}

We can also use the Fisher to consider the impact of weight perturbations on the predictions over the entire dataset as
\begin{align*}
    \DeltaKL(\dataset) = \frac{1}{\ndata} \sum_{i=1}^\ndata \DeltaKL(\x^{(i)}) 
    &= \frac{1}{\ndata} \sum_{i=1}^\ndata d\param^\T \FIM^{(i)}_{\param^*} d\param \\
    &= d\param^\T \underbrace{\left( \frac{1}{\ndata} \sum_{i=1}^\ndata \FIM^{(i)}_{\param^*} \right)}_{:= \FIM^\dataset_{\param^*}} d\param.
\end{align*}

\subsection{Connection to the Hessian}
As is evident from \eqref{eq:fisher_def}, there are strong connections between the dataset Fisher $\FIM^\dataset_{\param^*}$ and the Hessian with respect to $\param$ of the log likelihood of the training data. If we define $L(\param) = \sum_{i=1}^\ndata \log p_{\param^*}(\y^{(i)} \mid \x^{(i)})$,
the Hessian of $L$ evaluated at $\param^*$ can be well approximated by the dataset Fisher\footnote{By application of the chain rule, $H_L = \hat{\FIM}^\dataset_{\param^*}+ C(\param^*)$, where $\hat{\FIM}^\dataset_{\param^*}$ is the empirical Fisher where the expectation in \eqref{eq:fisher_def} is replaced by an empirical expectation over the dataset, and $C(\param^*)$ is a term involving first derivatives of the log likelihood and second derivatives of the network $f$. For trained networks, we expect the Fisher and the empirical Fisher to be closely aligned, and $C(\param^*) \approx 0$ since the first derivatives of the log likelihood are near zero at the end of training. Furthermore, for piece-wise linear networks (e.g., with ReLU activations), the second deriviatves of $f$ are 0, and so, $C(\param^*) = 0$.} \citep{Ritter2018ASL,martens2015optimizing}:
\[
    H_L = \frac{\partial^2 L}{\partial \param^2} \biggr|_{\param=\param^*} \approx \ndata \FIM^\dataset_{\param^*}.
\]
Unlike the Hessian, the Fisher is always guaranteed to be positive semidefinite. For this reason, this approximation is common in the field of second-order optimization, where preconditioning gradient steps with the inverse Fisher tends to be more efficient and numerically stable than using the Hessian.

\subsection{The Laplace Approximation of Epistemic Uncertainty}
The Hessian of the log-likelihood has strong connections to Bayesian ideas of \textit{epistemic uncertainty}, the uncertainty due to lack of data. From a Bayesian perspective, one can choose a prior over the weights of a DNN, $p(\param)$, and then reason about the posterior distribution on these weights given the dataset, $p(\param \mid \dataset)$. Often, due to the overparameterized nature of DNNs, many values of $\param$ are likely under this dataset, corresponding to different ways the DNN can fit the training data \citep{azizan2019stochastic}. By characterizing the posterior, and then marginalizing over it to produce a posterior predictive distribution $p(\y \mid \x) = \int p(\param \mid \dataset) p_{\param}(\y \mid \x) d\param$, one can hope to detect atypical data by incorporating uncertainty due to lack of data into the network's probabilistic predictions. 

While computing this posterior is intractable for DNNs, due to their nonlinearity and high-dimensional weight space, many approximations exist, leveraging Monte-Carlo sampling, or by assuming a distributional form and carrying out variational inference. One approximation is the Laplace approximation \citep{mackay1992practical}, which involves a second-order approximation of the log posterior $\log p(\param \mid \dataset)$ about a point estimate $\param^*$. This quadratic form yields a Gaussian posterior over the weights. If the prior on the weights is $p(\param) = \N(\cdot; 0, \epsilon^2 I_\nparam)$, the Laplace posterior is given by 
$\Sigma^* = \frac12 \left(H_L + \frac{1}{2\epsilon^2} I_\nparam \right)^{-1}$. 
The Laplace approximation is attractive as it uses local second-order information (the Hessian $H_L$) to produce an estimate of the Bayesian posterior for any pretrained model. However, even computing this approximation to the exact Bayesian posterior can be challenging for large models, where estimating and inverting an $\nparam \times \nparam$ matrix to compute $\Sigma^*$ is demanding.

%% file: 4_approach.tex
In this section, we describe the main steps in our OoD detection method.
We build upon ideas from the Laplace approximation, but using the dataset Fisher to approximate the Hessian. Thus, our approximate posterior is given by
\begin{align}
    \Sigma^* &= \frac12 \left(M \FIM^\dataset_{\param^*} + \frac{1}{2\epsilon^2} I \right)^{-1}. \label{eq:posterior-cov}
\end{align}
As we are only focused on the downstream-task OoD detection, we avoid the computationally expensive step of marginalizing over this uncertainty to form a posterior predictive distribution, and instead, directly quantify how this posterior uncertainty on the weights impacts the networks probabilistic predictions. 
A natural metric for that purpose is the expected change in the output distribution (measured by the KL divergence) when the weights are perturbed according to the Laplace posterior distribution, i.e.,
\begin{align}
    \oodmetric(\x^\test) &= \underset{d\param \sim \N(0, \Sigma^*)}{\E}\left[ \DeltaKL(\x^\test) \right] \nonumber \\
    &\approx \underset{d\param \sim \N(0, \Sigma^*)}{\E} \left[ d\param^\T \FIM^\test_{\param^*} d\param \right] \nonumber \\
    &= \trace\left( \FIM^\test_{\param^*} \Sigma^* \right). \label{eq:oodmetric_simple}
\end{align}
Crucially, by using the Fisher to estimate the change in the output distribution due to weight perturbation, we obtain a quadratic form whose expectation we can compute analytically. 
Note that this metric is the Frobenius inner product between the test input's Fisher and a regularized inverse of the dataset Fisher. Thus, we can view this metric as measuring novelty in the terms of DNN's curvature, i.e., measuring the abnormality of a test input $\x$ by comparing the curvature at this input against the curvature on training inputs.

\subsection{Efficiently computing the Uncertainty Metric}
While \eqref{eq:oodmetric_simple} is a clean expression for an uncertainty metric, its na\"ive computation is intractable for typical DNNs, as both $\Sigma^*$ and $\FIM^\test_{\dataset}$ are $\nparam \times \nparam$ matrices. To make this computation amenable for real-time operation, we exploit the fact that both matrices are low-rank. From their definitions, it follows that $\rank(\FIM^\test_{\param^*}) \le \ndistparam$, and thus $\rank(\FIM^\dataset_{\param^*}) \le \ndata\ndistparam$. The number of weights in a neural network $\nparam$ is always greater than the output dimension $\ndistparam$, and for large models and typical dataset sizes, often also greater than $\ndata\ndistparam$. 
Thus, we choose instead to express both Fisher matrices in factored forms
\begin{align}
    \FIM^\test_{\param^*} = L^{\test}_{\param^*} L^{\test\T}_{\param^*}, \qquad
    \FIM^\dataset_{\param^*} = U \diag(\vec{\lambda}) U^T,
\end{align}
where $L^\test_{\param^*} \in \R^{\nparam \times \ndistparam}, U \in \R^{\nparam \times \ndata\ndistparam}$, and $\vec{\lambda} \in \R^{\ndata\ndistparam}$. 
Note that if we write $\FIM_{\distparam}(f(\x^{\test}, \param^*)) = L^{\test}_{\distparam^*}L^{\test\T}_{\distparam^*}$, then from \eqref{eq:reparameterization}, we can see that $L_{\param^*}^\test = \Jac_{f,\param}^\T L^\test_{\distparam^*}$. For many common choices of parametric distributions, $L_{\distparam^*}^\test$ can be computed analytically. Furthermore, leveraging the linearity of the derivative, we can compute each row of $L^\test_{\param^*}$ efficiently via backpropagation (see Appendix~\ref{app:backprop-tricks} for details and examples for common distributions).

Given these factored forms, an application of the Woodbury matrix identity allows us to simplify the computation to
\begin{align}
\begin{split}
    &\oodmetric(\x^\test) = \\
    &~~\epsilon^2 \left\| L^\test_{\param^*} \right\|^2_\mathrm{F} -
    \epsilon^2\left\| \diag\left( \sqrt{ \frac{\vec{\lambda}}{\vec{\lambda} + \nicefrac{1}{\left(2 M \epsilon^2\right)}}} \right) U^\T L^\test_{\param^*} \right\|^2_\mathrm{F} .
\end{split}
\end{align}
A derivation is provided in Appendix \ref{app:metric-derivation}.

In this new form, the computation is split into computing the factor $L_{\param^*}^\test$, carrying out the matrix product with $U^\T$, and then computing Frobenius norms.
The main bottleneck in this procedure, both in terms of memory and computation, is the matrix multiplication with the $\nparam \times \ndata\ndistparam$ matrix $U$. However, several empirical analyses of neural network curvature have found that the Hessian and dataset Fisher $\FIM^\dataset_{\param^*}$ exhibit rapid spectral decay \citep{sagun2017empirical, madras_detecting_2019}).
Indeed, we see that if $\lambda_i << 1/(2M\epsilon^2)$, the corresponding element in the diagonal matrix tends to 0. Thus, we can approximate this computation by only considering the top $\subspacedim$ eigenvalues and eigenvectors of the dataset Fisher, drastically reducing memory and computation requirements at test time. Notably, making this low-rank approximation gives us a strict \textit{over-estimate} of the exact quantity, which is well-suited for safety-critical settings, where being under-confident is more desirable than being over-confident. An error bound is provided in Appendix~\ref{app:metric-derivation}.

\subsection{Tractably approximating the Dataset Fisher via Matrix Sketching}

In order to compute $\oodmetric(\x^\test)$ online, we require the top $\subspacedim$ eigenvalues and eigenvectors of $\FIM^\dataset_{\param^*}$. While this computation can happen offline, it is still intractable to carry out exactly for common DNN models and large datasets, since simply representing $\FIM^\dataset_{\param^*}$ exactly requires storing an $\ndata \ndistparam \times \nparam$ factor, which can easily grow beyond the capacity of common GPU memories for large perception networks and datasets with tens of thousands of parameters.
To alleviate this issue, prior work has considered imposing sparsity patterns on the Fisher, e.g., diagonal (which ignores correlations between weights), or layer-wise block-diagonal \citep{Ritter2018ASL} (which ignores correlations between layers).
Instead, we note that only the top eigenvectors of the datastet Fisher are important to the computation of our uncertainty metric, and thus we turn to tools from matrix sketching to tractably estimate a low-rank approximation of the Fisher \textit{without} imposing any sparsity structure on the matrix.

The key idea in matrix sketching, to avoid working with a large matrix directly, is to apply a randomized linear map $\mathsf{S}$ to the matrix of interest \citep{tropp_practical_2017}. By appropriately randomizing this map (the \textit{sketching operator}), we obtain high-probability guarantees that the image of the original matrix produced by the map (the \textit{sketch}) encodes sufficient information about the original matrix.
Given a bound on the desired approximation error, the size required for the sketch depends on the desired rank of the approximation $\subspacedim$, and not the original size of the large matrix. Thus, this can enable our technique to be applied to arbitrarily large datasets. 

The linearity of the sketching operator allows us to form this sketch iteratively, using a single pass over the dataset, without storing the full dataset Fisher in memory. 
Specifically, from the definition of $\FIM^\dataset_{\param^*}$, we can compute its sketch as the sum of smaller sketches:
\begin{align}
    \sketchop\left(\FIM^\dataset_{\param^*}\right) &= \frac1\ndata \sum_{i=1}^\ndata \sketchop\left( \FIM^{(i)}_{\param^*} \right) 
    = \frac1\ndata \sum_{i=1}^\ndata \sketchop\left( L^{(i)}_{\param^*} L^{(i)\T}_{\param^*} \right). \label{eq:total-sketch}
\end{align}

Following \cite{tropp_practical_2017}, we choose $\sketchop$ to independently left- and right-multiply the Fisher by random sketching matrices. Specifically, the sketch of each component is computed as $Y^{(i)},W^{(i)} \gets \sketchop( L^{(i)}_{\param^*} L^{(i)\T}_{\param^*} )$, where
\begin{align}
Y^{(i)} &= \left( \left(\Omega L^{(i)}_{\param^*} \right) L^{(i)\T}_{\param^*}\right)^\T, &
    W &= \left( \Psi L^{(i)}_{\param^*} \right) L^{(i)\T}_{\param^*}, 
\end{align}
where $\Omega \in \R^{r \times \nparam}$ and $\Psi \in \R^{s \times \nparam}$ are random sketching matrices, with $T = r+s$ defining the total size of the sketch. Note that following the operation order indicated by the parentheses avoids instantiating any $\nparam \times \nparam$ matrix. The memory complexity of this operation is $O(T(\nparam + \ndistparam))$. \citet{tropp_practical_2017} suggest splitting the budget to $r = (T-1)/3, s = T - r$, and suggest choosing $T=6\subspacedim+4$ as a minimal value of $T$ for a given $\subspacedim$ to minimize a high-probability bound on the approximation error of the sketch. We refer the reader to \citep{tropp_practical_2017} for a discussion of these theoretical results.

These sketching matrices can be as simple as matrices with i.i.d. standard Gaussian entries. However, to further reduce the memory and computation overhead of sketching, in this work, we use Subsampled Randomized Fourier Transform (SRFT) sketching matrices \citep{woolfe2008fast}
\begin{align}
    \Omega &= P_1 F_{\nparam} \diag(\vec{d}_1), & \Psi &= P_2 F_{\nparam} \diag(\vec{d}_2), \label{eq:sketch-map}
\end{align}
where $\vec{d}_1, \vec{d}_2 \in \R^\nparam$ are vectors with entries sampled from independent Rademacher random variables\footnote{A Rademacher random variable has a value of $+1$ or $-1$ with equal probability.}, $F_{\nparam}$ is the linear operator which applies the discrete cosine transform on each $\nparam$-length column, and $P_1, P_2$ are matrices which each select a random subset of $r$ and $s$ rows respectively. The SRFT sketching matrices offer similar approximation performance when compared to Gaussian matrices, but adding only a $(T+2\nparam)$ parameter overhead, as opposed to $T\nparam$ of the Gaussian case \citep{tropp_practical_2017}.

\begin{algorithm}[t]
\caption{\algName{} Offline\label{alg:offline}}
\begin{algorithmic}[1]
\Require{Dataset $\dataset = \{ \x^{(i)}, \y^{(i)} \}_{i=1}^\ndata$, DNN architecture $f, \mathcal{P}$, trained weights $\param^*$}
\Function{SketchCurvature}{$f, \mathcal{P}, \param^*, \dataset$}
\State Sample $\Omega, \Psi$ as in \eqref{eq:sketch-map}. \Comment{construct sketching map}
\State $Y, W \gets 0, 0$ \Comment{initialize sketch}
\For{$(\x^{(i)}, \y^{(i)})$ \textbf{in} $\dataset$}
    \State $\distparam^{(i)} \gets f(\x^{(i)}, \param^*)$ \Comment{forward pass}
    \State Compute $L^{(i)}_{\param^*}$ from $\distparam^{(i)}$ \Comment{$\ndistparam$ backward passes}
    \State $Y \gets Y + \frac{1}{\ndata} \left( \left(\Omega L^{(i)}_{\param^*} \right) L^{(i)\T}_{\param^*}\right)^\T$ \Comment{update sketch}
    \State $W \gets W + \frac{1}{\ndata} \left( \Psi L^{(i)}_{\param^*} \right) L^{(i)\T}_{\param^*} $
\EndFor
\State $\subspacebasis, \vec{\lambda}_\mathrm{top} \gets$ \Call{FixedRankSym}{$\Omega, \Psi, Y, W$}
\State \Return $\subspacebasis, \vec{\lambda}_\mathrm{top} $
\EndFunction
\end{algorithmic}
\end{algorithm}

Armed with these tools, we can now follow the procedure detailed in Algorithm \ref{alg:offline} to produce a low-rank approximation of the dataset Fisher.
Our approach uses one pass over the dataset $\dataset$, incrementally building the sum in \eqref{eq:total-sketch} by applying the SRFT sketching matrices to the $L^{(i)}_{\param^*}$, the factor for the Fisher for a single input, which can be computed with $\ndistparam$ backward passes for each input. Having constructed the sketch, we can use this much lower-dimensional, $T \times \nparam$ representation of the dataset Fisher to extract a low-rank, diagonalized representation $\FIM^{\dataset}_{\param^*} = \subspacebasis \diag(\vec{\lambda}_\mathrm{top}) \subspacebasis^\T$ by following the \texttt{FixedRankSymmetric} algorithm detailed in \citet{tropp_practical_2017}. 

%% file: 5_related_work.tex
There is a large body of work on characterizing epistemic uncertainty in DNNs. Specific to softmax-based classification models, there has been some effort in using the predictive uncertainty directly as a measure of epistemic uncertainty for OoD detection \citep{hendrycks_baseline_2018}, which can be improved through temperature scaling and input pre-processing \citep{liang_enhancing_2018}. Bayesian approaches often employ Monte-Carlo \citep{neal2012bayesian,gal2015dropout} or variational \citep{graves2011practical,blundell2015weight,liu2016stein} methods, but these are not generally applicable to pre-trained networks. In contrast, the Laplace approximation to the Bayesian posterior \citep{mackay1992practical} is appealing as it can be applied to any pre-trained DNN, yet requires estimating the network curvature.
    
The most closely-related approaches to ours are the post-training uncertainty methods of \citet{madras_detecting_2019} and \citet{Ritter2018ASL}.
        The main differences between such approaches are (1) how to approximate the curvature (Hessian/Fisher) and (2) how to propagate uncertainty at test time.
        The true Bayesian posterior is based on the log-likelihood, and a second-order approximation of the log-likelihood would involve the Hessian (the Laplace approximation).
        However, for DNNs, we often do not optimize to convergence, and thus, the Hessian is not guaranteed to be PSD. \citet{madras_detecting_2019} consider only the principal components, but the computation still does not scale well to large datasets. Moreover, stochastic versions of Lanczos algorithm \citep{lanczos1950iteration} or power iteration are difficult to tune.
        An alternative is to consider the Fisher (or the Gauss–Newton \citep{botev2017practical}) approximation of the full Hessian. For certain cases, such as for the exponential family distributions and piecewise linear networks, the Fisher is the same as the Hessian\footnote{For piecewise linear networks, e.g., those with ReLU activations, the Hessian of $\distparam$ with respect to the weights is 0 wherever the network is differentiable, and the nondifferentiable points of such networks form a measure zero set \citep{singla2019understanding}
        }.
        An advantage of the Fisher is that it is easier to compute, as it only requires the first-order terms.
        Further, it is guaranteed to be PSD, and thus is often used in second-order optimization. 
        
        As mentioned earlier, Fisher is still intractable to compute, store, and invert for large models, and so, various approximation schemes for that have been proposed. \citet{Ritter2018ASL} use a Kronecker-factored representation of the Fisher, which is easy to store and invert, but it requires certain approximations of the expectations of a Kronecker product, and ignores the cross terms between layers to form a block diagonal structure.
        We do not enforce this block diagonal structure, and instead use matrix sketching to tractably estimate only the top eigenvectors and eigenvalues of the full Fisher. 
        The KFAC approximation (and its derivatives) are suitable for second-order optimization, when curvature is used to scale the gradient step in the training loop (making the block diagonal approximation enables quick computation of an invertible Fisher). However, in our case, computing the dataset Fisher (and its eigen-decomposition) happens only once, offline, and thus, we can afford to take a slower, more memory-intensive approach.

    While proposed for a different purpose, there are similar ideas that have been used for the problem of continual learning, i.e., learning different tasks in a sequential fashion. The challenge there is to represent the information of the previous training data in a compact form and to update the weights in such a way that preserves the previous information as much as possible when training for a new task. Elastic weight consolidation (EWC) of \citet{kirkpatrick2017overcoming} uses the Fisher information matrix to weight each parameter based on its ``importance'' for the previous task, and uses a regularizer that penalizes changing important parameters more.  \citet{farajtabar2020orthogonal} proposed orthogonal gradient descent (OGD), which represents the information about the previous data in the form of gradients of the predictions, and then updates the weights in a direction orthogonal to those gradients.

%% file: 6_experiments.tex
We are interested in evaluating how \textit{efficiently} \algName{} produces uncertainty estimates, and how \textit{useful} these estimates are. We define the utility of the uncertainty estimate in terms of how well it serves to classify atypical inputs from typical ones. Following the literature on OoD detection, we quantify this utility by computing the area under the ROC and precision-recall curves, to produce the AUROC and AUPR metrics, respectively. 

\begin{figure}
    \centering
    \includegraphics[width=\columnwidth]{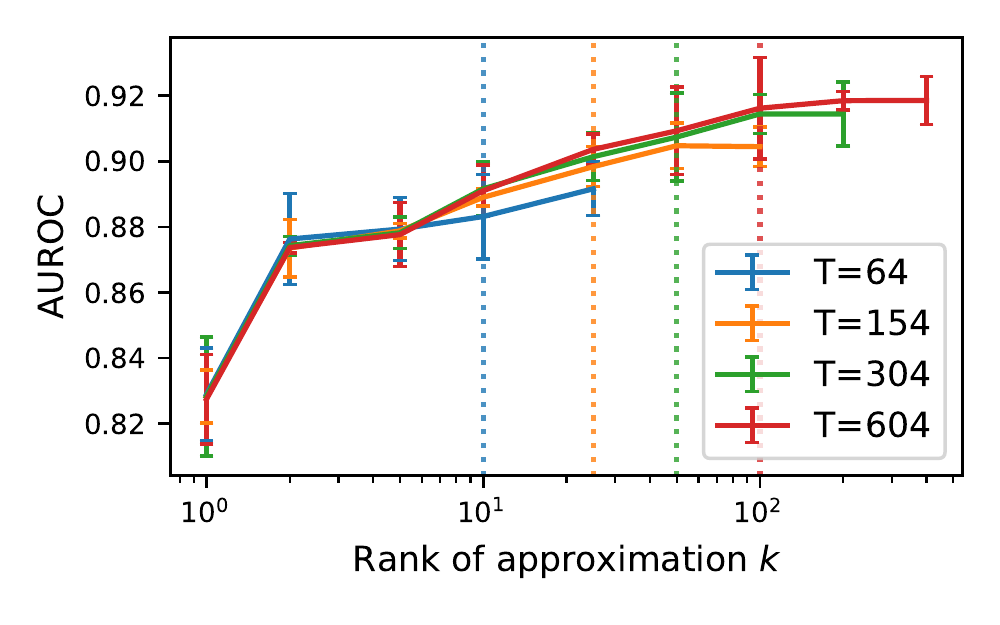}
    \caption{The impact of the sketched approximation on OoD detection (measured by AUROC) for Rotated MNIST. We see that, in general, increasing the rank of the approximation yields a higher AUROC, but with diminishing returns (note the log scale). AUROC is not significantly impacted by the sketch budget $T$ if $\subspacedim \le (T-4)/6$, the threshold visualized by the dashed lines.}
    \label{fig:auroc-vs-rank}
\end{figure}
Using these metrics, we explore: (1) how choices like the sketch budget $T$ and rank of approximation $\subspacedim$ impact performance; (2) how performance of \algName{} can be improved on large DNNs; and (3) how \algName{} compares to baselines on a suite of problem settings, from regression to classification.
The details of all the problem settings and the experimental evaluation are included in Appendix \ref{app:domains}.

\begin{figure*}
    \centering
    \includegraphics[width=\textwidth]{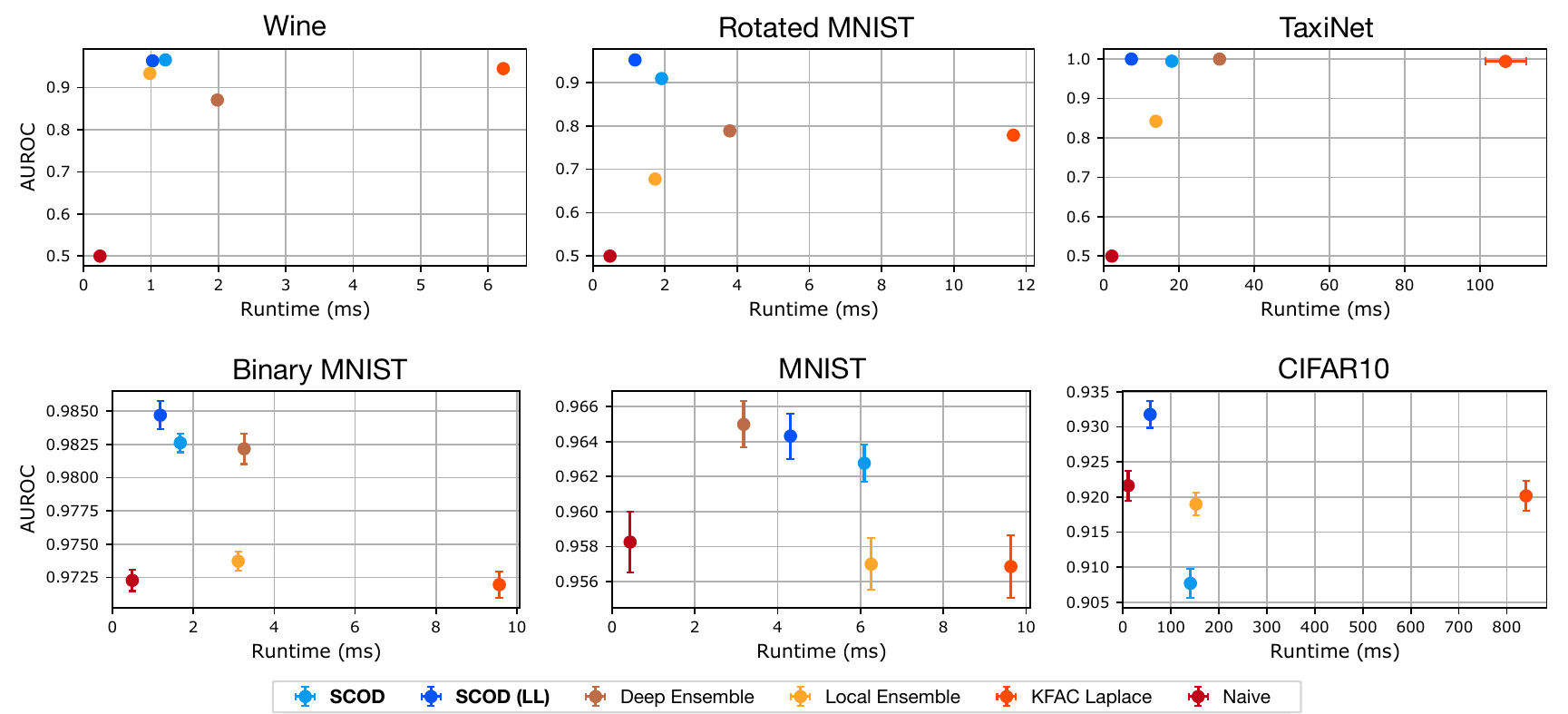}
    \caption{Comparing OoD detection performance and runtime against baselines. We see that \algName{} is consistently on a Pareto frontier in terms of the runtime/efficacy tradeoff among post-hoc uncertainty methods (top-left is better). Indeed, \algName{}, applied onto a pre-trained model, often matches or exceeds the performance of Deep Ensembles, which is not a post-training approach, and requires retraining multiple models from scratch. Note that the y-axes are independently scaled.}
    \label{fig:runtime-vs-auroc}
\end{figure*}

\subsection{Choosing the Sketch Budget and Rank of Approximation}
A key aspect of \algName{} is using matrix sketching to approximate the dataset Fisher as a low-rank matrix. This presents the practitioner with two key hyperparameters: the memory budget $T$ to allocate for the sketching, and the rank $\subspacedim$ of the approximation used in online computation. While memory budget $T$ is generally set by hardware constraints, it impacts the quality of low-rank approximation attainable through sketching. Indeed, the sketching procedure produces a rank $2(T-1)/3$ approximation, and theoretical results suggest keeping only the top $(T-4)/6$ eigenvalues and eigenvectors from this approximation \citep{tropp_practical_2017}. 

We explore this trade-off empirically on the Rotated MNIST domain. We choose a range of values for the sketching budget $T$, and sketch the dataset Fisher. For each sketch size, we choose $\subspacedim$ from a range of values from $1$ all the way to the theoretical maximum $2(T-1)/3$, and test the AUROC performance of $\oodmetric$. Figure \ref{fig:auroc-vs-rank} shows the results of these experiments.
These results show two key trends. First, we see that increasing the rank $\subspacedim$ improves performance, but with diminishing returns. Second, for a fixed rank $\subspacedim$, we see that the performance is insensitive to the sketch budget, especially if $\subspacedim  < (T-4)/6$. For larger $\subspacedim$, we start seeing benefits from increasing the sketch budget, consistent with the theory; performance starts to plateau as $\subspacedim$ increases beyond $(T-4)/6$. 
Beyond the rank, $\oodmetric$ is also impacted by the scale of the prior $\epsilon^2$. In our experiments, we found performance to be insensitive to this hyperparameter and thus use $\epsilon=1$ in all the experiments. See Appendix \ref{app:Meps-sweep} for more details.

\subsection{Further improving efficiency for large DNNs}
These results suggest that the best classification performance is achieved by setting $T$ as high as memory allows, and subsequently choosing $\subspacedim \ge (T-4)/6$. While sketching enables us to avoid the quadratic dependency on $\nparam$, the memory footprint of the offline stage of \algName{} is still linear, i.e., $O(\nparam T)$. As a result, GPU memory constraints can restrict $T$, and thus $\subspacedim$, substantially for large models. To alleviate this issue, we study how performance is impacted if we simply restrict our analysis to the last few layers of the network, thus lowering the effective value of $\nparam$. For convolutional networks, the first layers tend to learn generic filter patterns which apply across many domains, while later layers learn task-specific representations \citep{zeiler2014visualizing}. Thus, it is possible that the curvature on the later-layer parameters is more informative from an OoD detection standpoint. 

To test the impact of this, we compare to \textbf{\algName{} (LL)}, an ablation which applies this analysis only to the last layers of the network. For these experiments, we chose to limit analysis to the last $L$ layers, where we chose $L$ for each network architecture such that a minimum of 2 layers were considered, and at least 1 Conv layer was considered. For the larger models in TaxiNet and CIFAR10, we restricted our analysis to the last 15\% of the layers. In these latter two domains, we also increased the sketch budget $T$ and associated rank $\subspacedim$ up to the capacity of our GPU. Full details of the setup are included in Appendix \ref{app:domains}. Results are included in Figure \ref{fig:runtime-vs-auroc}. For an in-depth look into the impact of focusing on the last layer, see Appendix \ref{app:last-layers}.

\subsection{Performance relative to Baselines}
We compare \algName{} against several baselines.
First, we compare with the two closely-related approaches, namely, \textbf{Local Ensembles} \citep{madras_detecting_2019} and \textbf{KFAC Laplace} approximation \citep{Ritter2018ASL}, which use curvature estimation to augment a trained model with uncertainty estimates.
Next, while \textit{not} a method applicable to a pre-trained model, we compare against \textbf{Deep Ensembles} \citep{lakshminarayanan2017simple} as a benchmark, as it has shown strong OoD detection performance across regression and classification examples. For this baseline, we retrain $K=5$ models of the same architecture from different initializations.
Both KFAC Laplace and Deep Ensembles output mixture distributions, which we turn into a scalar uncertainty measure by computing the total variance (summed over output dimension) for regression problems, or the predictive entropy for classification, as in \citep{lakshminarayanan2017simple}.
Finally, we compare to a \textbf{Naive} baseline which produces an uncertainty measure directly from the output of the pre-trained model. For regression models which output only a mean estimate, extracting an uncertainty estimate is not possible, and thus this baseline outputs a constant signal $\oodmetric(x) = 1$. For classification models, we use the entropy of the output distribution as a measure of uncertainty, as in \citep{lakshminarayanan2017simple}.

We compare the performance of these baselines across three regression domains (Wine, Rotated MNIST, and TaxiNet), as well as three classification domains (Binary MNIST, MNIST, and CIFAR10). 
For each domain, we create a dataset known to be semantically different from the training data. Where possible, we consider OoD data that is realistic --- for example, data from white wines as OoD for a model trained on data form red wines; or in TaxiNet, images from a wing-mounted camera in different times of day and different weather conditions for a model trained only on data from the morning with clear weather conditions. We also consider using the model's own accuracy to label points as in- or out-of-distribution in Appendix \ref{app:error-based}. For all domains except CIFAR10, we train all networks, and then apply the post-training algorithms. For the CIFAR10 domain, we test on a pre-trained DenseNet121 model \citep{huang2019convolutional, huy_phan_2021_4431043} to highlight the fact that \algName{} can be applied to augment any pre-trained model, independent of training methodology, with uncertainty estimates.

The results are summarized in Figure \ref{fig:runtime-vs-auroc}. Overall, we see three key trends. First, \algName{} consistently provides the most informative uncertainty measures out of the methods applicable on a pre-trained model. In fact, its AUROC often matches or exceeds that of Deep Ensembles. While KFAC Laplace also matches Ensemble performance in many settings, our approach tends to dominate it on a runtime/AUROC Pareto frontier. Computing uncertainty metrics using KFAC Laplace requires the computationally-intensive repeated evaluations of the DNN with different sampled weights. Furthermore, this sampling process is very sensitive to the regularization hyperparameters. In contrast, our uncertainty metric does not require any sampling, and thus adds very little overhead. Local Ensembles similarly add little overhead, but have a worse AUROC on many domains, especially for large models and datasets, e.g., in the TaxiNet domain, where $\ndata = 50,000$. It is likely that sketching the Fisher yields a better low-rank curvature estimate than using stochastic mini-batching to estimate the top eigenspace of the Hessian.

On regression problems, the output of the network provides no estimate of uncertainty, and, unsurprisingly, the Naive baseline performs poorly. In contrast, on classification problems, the Naive strategy of interpreting the predictive uncertainty as epistemic uncertainty works quite well \citep{hendrycks_baseline_2018}. Nevertheless, we see that, \algName{} is generally able to exceed the Naive performance on these classification domains, and match the performance of Deep Ensembles. Importantly, unlike Deep Ensembles or even KFAC Laplace, we do not directly use the base DNN's output uncertainty as part of our score, and only consider how weight perturbations may change the output. Given that the output uncertainty is a strong baseline for softmax classification problems, connecting these ideas is an interesting avenue for future work. 

In general, we see that restricting analysis to the last layers often leads to better AUROC performance as well as faster runtime. This is particularly evident on CIFAR10, where without restricting analysis to the last layers, \algName{} performed worse than the Naive baseline (although still achieving AUROC > 0.9). We suspect this is due to the first layers representing generic features that are suited to most natural images, and thus, the curvature of the later layers is more useful for OoD detection. This is supported by a case study on the CIFAR10 model, with results in Appendix \ref{app:last-layers}.

%% file: 7_conclusions.tex
We presented \emph{\algNameLong{} (\algName{})}, a scalable,  architecture-agnostic framework for equipping any pre-trained DNN with a task-relevant epistemic uncertainty estimate. Through extensive experiments, we demonstrated that the proposed method achieves comparable or better OoD detection performance with a lower computational burden relative to existing approaches.

There are several important avenues for future work, in addition to what was mentioned earlier. Our framework can be potentially extended to continual learning settings, since the sketch can be built incrementally without the need to loop over the previous data. Furthermore, within the context of autonomous systems, another important direction is to tailor OoD detection and uncertainty estimation for the downstream control task.

%% file: A_appendix.tex
\section{Computing per-input Fisher via backprop}
\label{app:backprop-tricks}
We work with the factorization $\FIM^\test_{\param^*} = L^\test_{\param^*}L^{\test\T}_{\param^*}$
Recall that $L^{\test}_{\param^*} = L_{\param^*}^\test = \Jac_{f,\param}^\T L^\test_{\distparam^*}$.
Leveraging the linearity of the derivative, we can avoid carrying out this matrix multiplication on $\Jac^{\test}_{f,\param^*} \in \R^{\ndistparam \times \nparam}$, and instead perform the matrix multiplication prior to computing the Jacobian via backpropagation. Defining the function $\vec{g}(A, \x, \param) = A f(\x,\param)$, which applies a linear transformation $A$ to the output of the DNN, we have 
\begin{align}
    L^{\test\T}_{\param^*} &= \left[ \begin{array}{c}
        \frac{\partial}{\partial\param} g_1 (L^{\test\T}_{\distparam^*}, \x^\test, \param^*) \\
        \vdots \\
        \frac{\partial}{\partial\param} g_\ndistparam (L^{\test\T}_{\distparam^*}, \x^\test, \param^*)
    \end{array} \right],
\end{align}
where each row can be computed by backpropagation from the corresponding output dimension of $\tilde{\distparam}$, but ignoring the gradients that flow through the dependence of $L^{\test\T}_{\distparam^*}$ on $\param$.

Alternatively, for models with large output dimensions, exactly computing the Fisher as outlined above can be expensive. For such settings, it is possible to exploit the definition of the Fisher as an expectation over $\y \sim P(\distparam)$, and turn to numerical integration techniques such as Monte-Carlo estimation. In our experiments, we use the exact Fisher in both the offline and online phases.

Below, we provide analytic forms of $L^\test_{\distparam^*}$ for common parameteric distributions:
\begin{itemize}
    \item \textbf{Fixed Diagonal Variance Gaussian.} If $\mathcal{P}(\distparam)= \N(\distparam, \diag(\vec{\sigma}))$, then 
    \[
        L^\test_{\distparam^*} = \diag(\vec{\sigma})^{-\nicefrac12}.
    \]
    \item \textbf{Bernoulli with Logit Parameter.} If the output distribution is chosen to be a Bernoulli parameterized by the logit $\distparam \in \R$, such that the probability of a positive outcome is $p = 1/(1+\exp(-\distparam^*))$, then, $\FIM^\test_{\distparam^*} = p(1-p)$. So,
    \[
        L^\test_{\distparam^*} = \sqrt{p(1-p)}.
    \]
    \item \textbf{Categorical with Logits.} If the output distribution is a categorical one parameterized by the logits $\distparam \in \R^\ndistparam$, such that $p(y = k) = \frac{\exp(\distparam_k)}{\sum_{j=1}^\ndistparam \exp(\distparam_j)}$, then,
    \[
        L^\test_{\distparam^*} = \diag(\vec{p})^{\nicefrac12} (I_\ndistparam - \vec{1}_{\ndistparam} \vec{p}^\T),
    \]
    where $\vec{p}$ is the vector of class probabilities according to $\distparam^*$.

\end{itemize}

\section{Derivation of the Simplified Uncertainty Metric}
\label{app:metric-derivation}

If we substitute the eigenvalue decomposition of the dataset Fisher $\FIM^\dataset_{\param^*} = U \diag(\vec{\lambda}) U^T$ into the expression for the posterior covariance \eqref{eq:posterior-cov}, and apply the Woodbury identity, we obtain
\begin{align}
    \Sigma^* &= \epsilon^2 \left( I - U \diag\left( \frac{\vec{\lambda}}{\vec{\lambda} + \nicefrac{1}{\left( 2 M \epsilon^2 \right)}} \right) U^\T \right),
\end{align}
where the operations in the diagonal are applied elementwise.

Now, if we plug this expression into our expression for $\oodmetric$, we obtain
\begin{align}
    &\oodmetric(\x^\test)\notag\\
    &= \trace \left( \epsilon^2 \left( I - U \diag\left( \frac{\vec{\lambda}}{\vec{\lambda} + \nicefrac{1}{\left( 2 M \epsilon^2 \right)}} \right) U^\T \right) \FIM^\test_{\param^*} \right) \\
    &= \epsilon^2 \trace\left( \FIM^\test_{\param^*} \right) - \epsilon^2 \trace\left( U \diag\left( \frac{\vec{\lambda}}{\vec{\lambda} + \nicefrac{1}{\left( 2 M \epsilon^2 \right)}} \right) U^\T \FIM^\test_{\param^*} \right) \\
    &= \epsilon^2 \left\| L^\test_{\param^*} \right\|^2_\mathrm{F} - \epsilon^2 \trace\left( L^{\test\T}_{\param^*} U \diag\left( \frac{\vec{\lambda}}{\vec{\lambda} + \nicefrac{1}{\left( 2 M \epsilon^2 \right)}} \right) U^\T L^\test_{\param^*} \right) \\
    &= \epsilon^2 \left\| L^\test_{\param^*} \right\|^2_\mathrm{F} - \epsilon^2 \left\| \diag\left(\sqrt{ \frac{\vec{\lambda}}{\vec{\lambda} + \nicefrac{1}{\left( 2 M \epsilon^2 \right)}}} \right) U^\T L^\test_{\param^*} \right\|^2_\mathrm{F},
\end{align}
where we make use of the cyclic property of the trace and the fact that $\| A \|_F = \sqrt{\trace\left(AA^\T\right)}$.

\subsection{Low-rank Approximation and Error Bounds}
We notice that the elements of the diagonal $\frac{\lambda_j}{\lambda_j + \nicefrac{1}{2 M \epsilon^2}}$ tend to 1 for $\lambda_i >> \nicefrac{1}{2 M \epsilon^2}$, and 0 for $\lambda_i << \nicefrac{1}{2 M \epsilon^2}$. Therefore, only the top eigenvectors of the dataset Fisher are relevant to this posterior. We see that assuming a fixed spectral decay rate of $\FIM^\dataset_{\param^*}$, more eigenvectors are relevant if we choose $\epsilon^2$ to be large (wider prior weights), or $M$ to be large (more data points collected). The number of eigenvectors we keep influences memory and compute requirements. So, alternatively, we can choose a fixed rank of approximation $\subspacedim$, and then choose $\epsilon^2$ as appropriate.

Thus, we see that the dataset Fisher characterizes how the weights of the DNN are determined by the dataset. In fact, we see that this posterior distribution on the weights has a wide variance in all directions expect in the directions of the top eigenvectors of $\FIM^\dataset_{\param^*}$, for which $\lambda_j / \left(\lambda_j + (2 M \epsilon^2)^{-1}\right)$ is non-negligible. 

We can characterize the error made by keeping only the top $\subspacedim$ eigenvalues and eigenvectors. Let $\vec{\lambda}^\T = [\vec{\lambda}^\T_\mathrm{top}, \vec{\lambda}^\T_\mathrm{bot}], U = [\subspacebasis U_\mathrm{bottom}]$, where $\mathrm{top}$ selects the top $\subspacedim$ eigenvalues. If we define $\tilde{\oodmetric}(\x)$ as using the low-rank approximation, we have
\begin{align*}
    &\tilde{\oodmetric}(\x^\test)\\
    &= \epsilon^2 \left\| L^\test_{\param^*} \right\|^2_\mathrm{F} - \epsilon^2 \left\| \diag\left(\sqrt{ \frac{\vec{\lambda}_\mathrm{top}}{\vec{\lambda}_\mathrm{top} + \nicefrac{1}{\left( 2 M \epsilon^2 \right)}}} \right) \subspacebasis^\T L^\test_{\param^*} \right\|^2_\mathrm{F}.
\end{align*}

The error in the approximation can then be characterized as
\begin{align}
    &\tilde{\oodmetric}(\x^\test) - \oodmetric(\x^\test)\notag\\
    &= \epsilon^2 \left\| \diag\left(\sqrt{ \frac{\vec{\lambda}_\mathrm{bottom}}{\vec{\lambda}_\mathrm{bottom} + \nicefrac{1}{\left( 2 M \epsilon^2 \right)}}} \right) U^\T_\mathrm{bottom} L^\test_{\param^*} \right\|^2_\mathrm{F} \\
    &= \epsilon^2 \sum_{j=\subspacedim+1}^{\rank\left(\FIM^\dataset_{\param^*}\right)} \frac{\lambda_j}{\lambda_j + \nicefrac{1}{\left( 2 M \epsilon^2 \right) }} \left\| L^{\test\T}_{\param^*} \vec{u}_j \right\|_2^2 \\
    &\le \epsilon^2 \left\| L^{\test\T}_{\param^*} \right\|^2_2 \sum_{j=\subspacedim+1}^{\rank\left(\FIM^\dataset_{\param^*}\right)} \frac{\lambda_j}{\lambda_j + \nicefrac{1}{\left( 2 M \epsilon^2 \right) }}\\
    &\le \epsilon^2 \left\| L^{\test}_{\param^*} \right\|^2_F \sum_{j=\subspacedim+1}^{\rank\left(\FIM^\dataset_{\param^*}\right)} \frac{\lambda_j}{\lambda_j + \nicefrac{1}{\left( 2 M \epsilon^2 \right) }}\\
    &\le \epsilon^2 \left\| L^{\test}_{\param^*} \right\|^2_F \left({\rank\left(\FIM^\dataset_{\param^*}\right)}-k\right) \frac{\lambda_k}{\lambda_k + \nicefrac{1}{\left( 2 M \epsilon^2 \right) }}.
\end{align}

\section{Further Experimental Discussion}

\begin{figure}
    \centering
    \includegraphics[width=\columnwidth]{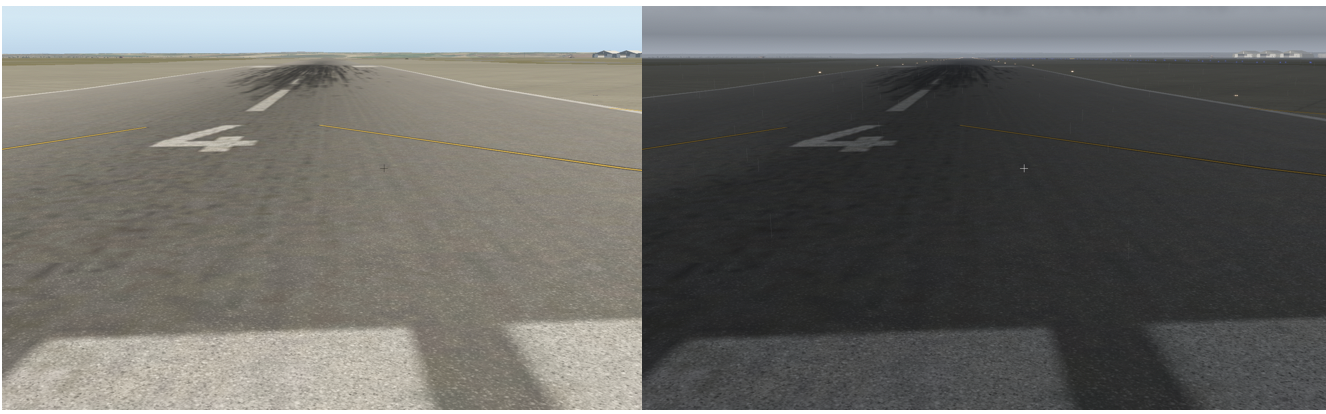}
    \caption{Example input images from the TaxiNet domain, in clear morning (left), and cloudy evening (right) conditions.}
    \label{fig:taxinet}
\end{figure}

\input{results_table}
\subsection{Experimental Domains}
\label{app:domains}
We evaluate \algName{} on several problem settings ranging from classification to regression. Here, we provide implementation details for all the domains. For all results, we report 95\% confidence bounds on AUROC and AUPR computed by bootstrapping. We measure performance on a system with an AMD Ryzen 9 3950X 16-core CPU and an NVIDIA GeForce RTX 2070 GPU.

Within regression, we consider the following datasets.
\begin{itemize}
    \item \textbf{Wine} is a dataset from the UCI Machine Learning Repository, in which inputs are various chemical properties of a wine, and labels are a scalar quality score of the wine. We train on the dataset of red wine quality and use the white wine dataset as OoD. The network architecture is a 3-layer fully connected network with ReLU activations, and each hidden layer having 100 units, yielding $\nparam = 11401$. Here
    $
    T = 604,
    k = 100.
    $ for \algName{}. For Local Ensembles, we use $k=20$ using all $\ndata=1000$ datapoints to compute exact Hessian-vector products.
    \item \textbf{Rotated MNIST}, where the input is an MNIST digit 2, rotated by a certain angle, and the regression target is the rotated angle. We consider two types of OoD data: the digit 2 rotated by angles outside the range seen at train time, as well as inputs that are other digits from MNIST. The model starts with three conv layers with $3\times3$ filters with a stride of 1. The number of channels in the conv layers is 16,32, and 32, with MaxPooling with a kernel size of 2 between each conv layer. The result is flattened and then processed by a linear layer with hidden dimension of 10 before the linear output layer, yielding a total of $\nparam = 16949$. We use ReLU activations. Here we use 
    \(
    T = 304,
    k = 50
    \) for \algName{}. For Local Ensembles, we use $k=50$ eigenvectors and all $\ndata=5000$ datapoints to compute exact Hessian-vector products. 
    \item \textbf{TaxiNet} which is a network architecture designed to process $3\times260\times300$ RGB input images from a wing-mounted camera and produce estimates of the aircraft's distance in meters from the centerline of the runway as well as its heading in radians relative to the runway, both of which can be used for downstream control during taxiing. It was developed by Boeing as part of the DARPA Assured Autonomy program\footnote{\url{https://www.darpa.mil/program/assured-autonomy}}. The model is based on a ResNet18 backbone, pre-trained on ImageNet, with the last layer replaced with a linear layer to the 2 output dimensions. This yields a total of $\nparam = 11177538$ weights. We fine-tune the network on data collected in the X-Plane 11\footnote{\url{https://www.x-plane.com/}} flight simulator with clear weather and at 9am. Here, we tested against realistic OoD data, by changing weather conditions to cloudy and changing the time-of-day to the afternoon and evening, which change the degree to which shadows impact the scene. Figure \ref{fig:taxinet} visualizes inputs under different conditions. Here $\ndata=30,000$. \algName{} processes all 30,000 datapoints in its sketch in under 30 mins. 
    Here,
    \(
    T = 46,
    k = 7
    \), for \algName{}, and for \algName{} LL, 
    \(
    T = 124,
    k = 20
    \). For Local Ensembles, we use $k=14$ eigenvectors of the Hessian.
\end{itemize}
For classification, we consider:
\begin{itemize}
    \item \textbf{BinaryMNIST}. We consider a binary classification problem created by keeping only the digits 0 and 1 from MNIST. As OoD data, we consider other MNIST digits, as well as FashionMNIST \cite{xiao2017/online}, a dataset of images of clothing that is compatible with an MNIST architecture. The network architecture we use has a convolutional backbone identical to that in the Rotated MNIST, with the flattened output of the conv layers processed directly by a linear output layer to a scalar output, for a total of $\nparam = 14337$ parameters. This output $\distparam$ is interpreted to be pre-sigmoid activation for logistic regression; i.e. the output parametric distribution is a Bernoulli with the probability of success given by $\mathrm{sigmoid}(\distparam)$. Here, we use
    \(
    T = 304,
    k = 50
    \) for \algName{}. For Local Ensembles, we use $k=50$ eigenvectors of the Hessian, computed with exact Hessian-vector products using all $\ndata=5000$ datapoints.
    \item \textbf{MNIST}. We also consider a categorical classification example formulated on MNIST, this time interpreting the output of the network as logits mapped via a softmax to class probabilities. We train on 5-way classification on the digits 0-4. We use digits 5-9 as OoD data, along with FashionMNIST. The network architecture is identical to that of BinaryMNIST, except here the output $\distparam \in \R^5$ represents the logits such that the probability of each class is given by $\mathrm{softmax}(\distparam)$. This yields $\nparam = 15493$. Here, we use
    \(
    T = 604,
    k = 100
    \) for \algName{}. For Local Ensembles, we use $k=100$ eigenvectors.
    \item \textbf{CIFAR10}. To test on larger, more realistic inputs, we consider the CIFAR-10 dataset \citep{krizhevsky2009learning}. Here, unlike the previous experiments, we use a pre-trained DenseNet121 model from \cite{huy_phan_2021_4431043}, and do not train the model from scratch, to highlight that \algName{} can be applied to any pre-trained model. While this model is trained on all 10 classes of CIFAR-10, we use only the first 5, and thus keep only the first 5 outputs as $\distparam \in \R^5$, and use them as the pre-softmax logits. This yields a total of $\nparam = 6956426$. We process $\ndata = 5000$ images sampled from the first 5 classes of the train split of CIFAR10, and use the val split as in-distribution examples to test on. For the experiments in the body of the paper, we use as the out-of-distribution dataset TinyImageNet\footnote{\url{http://cs231n.stanford.edu/tiny-imagenet-200.zip}}, a scaled down version of the ImageNet \citep{imagenet_cvpr09} dataset, keeping only 200 classes, and resizing inputs to $32\times32$ RGB images. In the tests in Appendix \ref{app:last-layers}, we also consider OoD data from the 5 held-out classes from CIFAR-10, as well as from the Street View House Numbers dataset \cite{Netzer2011} which has a compatible size. 
    Here, we use
    \(
    T = 76,
    k = 12
    \) for \algName{}, and 
    \(
    T = 184,
    k = 30
    \) for \algName{} (LL). For Local Ensembles, we use $k=20$ eigenvectors.
\end{itemize}

\subsection{Baselines}
We outline details of the implementation of each baseline here:
\begin{itemize}
    \item \textbf{Local Ensembles.} We reimplement this baseline in PyTorch, using the \texttt{pytorch-hessian-eigenthings} library \citep{hessian-eigenthings} to compute the top eigenspace of the Hessian. We implement Local Ensembles using a stochastic minibatch estimator for the Hessian-vector product for all domains except for Wine and the MNIST-based domains, where the models are small enough that using the full dataset to compute the Hessian vector product remains computationally feasible. At test time, we use the prediction gradient to compute the extrapolation score, while for multivariate regression and classification problems, we use the loss gradient, sampling possible outputs, computing the gradient on each, projecting it, and then aggregating by taking the minimum over the resulting scores.
    \item \textbf{KFAC Laplace.} We use the KFAC Laplace implementation found at \url{https://github.com/DLR-RM/curvature}. We use $30$ samples from the posterior prediction to estimate the Fisher, on a batch size of $32$. We found that in many cases, choosing default values for the norm and scale hyperparameters would lead to singular matrices making. For each experiment, we performed a coarse sweep over these hyperparameters, and chose the best performing set on a validation set to use for the results. Indeed, especially on the classification examples, we found that we required large values of the norm parameter to obtain accurate predictions, which regularizes the posterior towards a delta distribution centered around $\param^*$. 
    \item \textbf{Deep Ensemble.} We train $K=5$ models of identical architecture from random initializations using SGD on the same dataset. In all domains where Deep Ensembles are used, the first member of the ensemble is the same model as in the post-training  and Naive approaches. 
\end{itemize}

\begin{figure}
    \centering
    \includegraphics[width=\columnwidth]{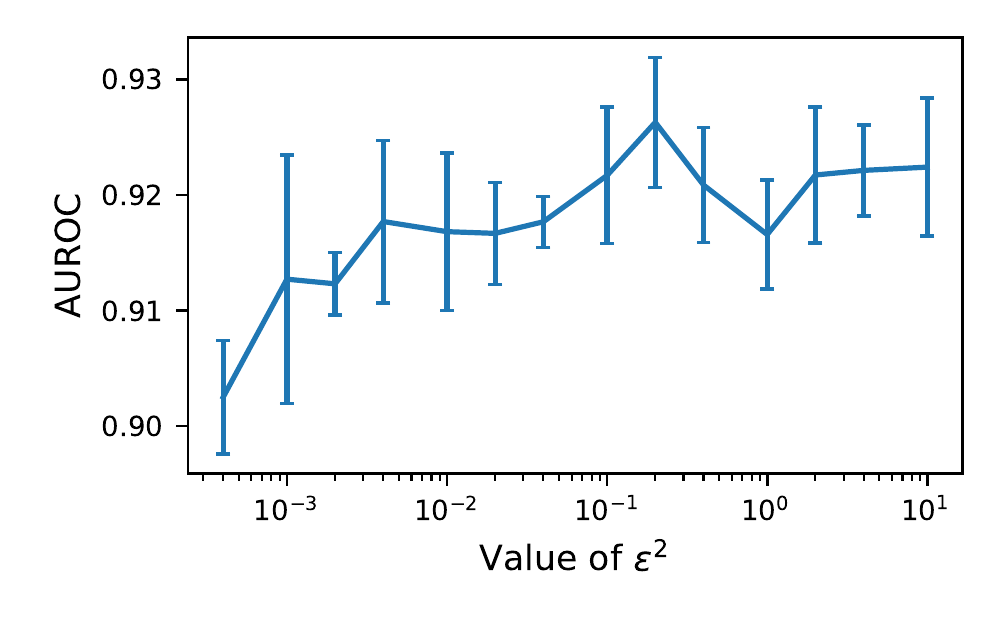}
    \caption{AUROC on RotatedMNIST as a function of prior scale $\epsilon^2$.}
    \label{fig:Meps-sweep}
\end{figure}

\subsection{Sensitivity to Scale of Prior}

\label{app:Meps-sweep}
We test the impact of the prior scale $\epsilon^2$ on the performance of \algName{} by performing a sweep on the Rotated MNIST domain, with $T=604, \subspacedim=100$. The results, shown in Figure \ref{fig:Meps-sweep}, suggest that this has little impact on the performance, unless it is set to be very small corresponding to very tight prior on the weights. This is unsurprising, as, apart from linearly changing the scale of our metric, the term only enters in the diagonal matrix, and has a significant impact if $1/(2\ndata\epsilon^2)$ is of comparable magnitude to the eigenvalues of the Fisher. This is rarely the case for the first $\subspacedim$ eigenvalues of the Fisher, especially when $\ndata$ is large (here, $\ndata = 5000$).

\begin{figure}
    \centering
    \includegraphics[width=\columnwidth]{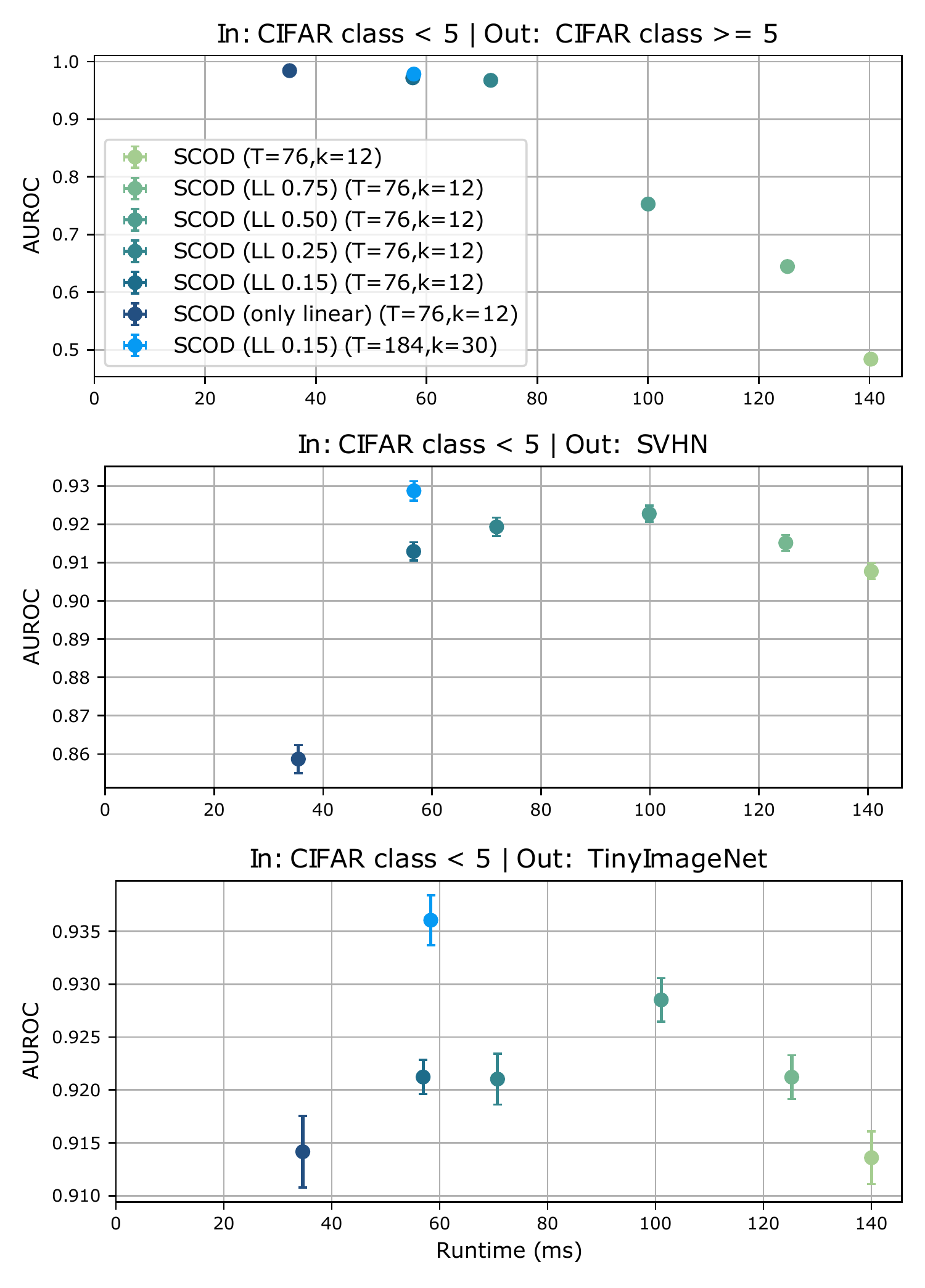}
    \caption{Impact of restricting analysis on the last layers of a network. We see that depending on the OoD dataset, there is a different optimal choice for which layers to consider.}
    \label{fig:last-layers}
\end{figure}

\subsection{Limiting \algName{} to the Last Layers}
\label{app:last-layers}
To explore the impact of only considering the last layers of a network, we consider restricting \algName{} to different subsets of the layers of the DenseNet121 model used in the CIFAR experiments. We hold the sketching parameters constant, and only vary how much of the network we consider. We denote this by the fraction after LL. For example, LL 0.5 means we restrict \algName{} to the last 50\%  of the network. With this notation, LL 1.0 represents considering the full network, which is simply \algName{}. At the other extreme, we consider restricting our analysis to just the last linear layer of the model, which we denote ``\algName{} (only linear).'' We consider performance with several different OoD datasets. Figure \ref{fig:last-layers} shows the results. The first OoD dataset is simply the other classes of CIFAR10 which we did not use to create the sketch. We see that here, using just the last linear layer performs best. Indeed, as the network was trained on all 10 classes, it is not surprising that only the last layer analysis provides any meaningful separation between the classes. Including the earlier layers ``dilutes'' this signal, and, for a fixed sketch budget, considering more layers leads to worse performance. On SVHN and TinyImageNet, we see that the optimal choice of the layers to consider is somewhere in between the two extremes. Here, it is possible that as we increase the fraction of the network that we analyze, we first gain the benefits of the information stored in the last layers, and then start to suffer the consequences of approximation error in the sketching, which depends on the size of the original matrix. We also visualize on this plot the performance of the version of \algName{} (LL) used to produce the results in the body of the paper. As is evident, we did not use the results of this sweep to optimally choose the number of layers to consider for the experiments in the body.

\begin{figure}
    \centering
    \includegraphics[width=\columnwidth]{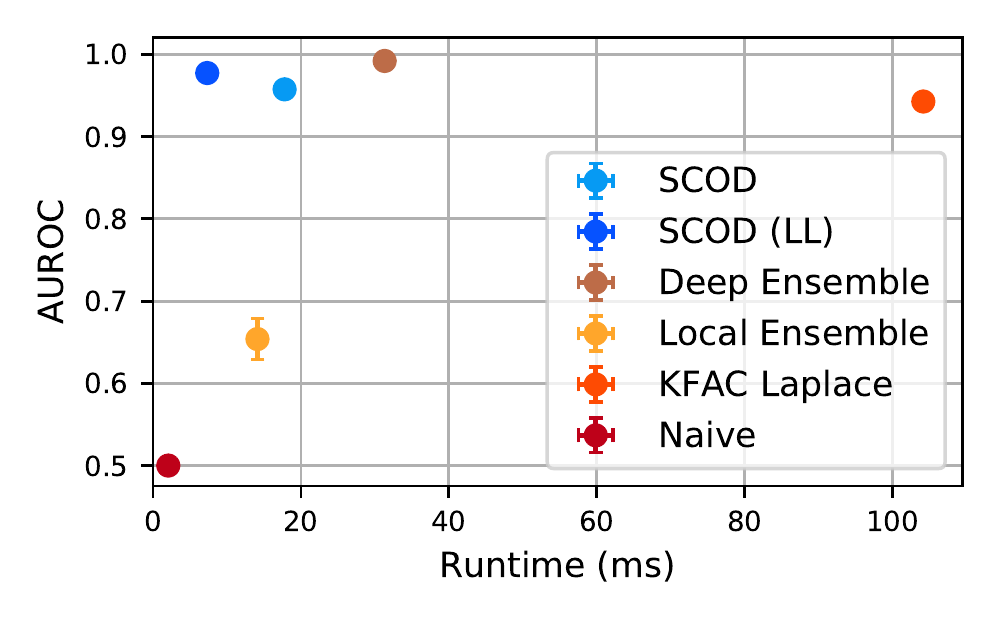}
    \caption{Error-based OoD detection on TaxiNet, where inputs where the DNN made an error greater than 0.5 in Mahalanobis distance are deemed to be out-of-distribution, i.e., out of the DNN's domain of competency.}
    \label{fig:error-based}
\end{figure}

\subsection{Error-based OoD detection}
\label{app:error-based}
In Figure \ref{fig:runtime-vs-auroc}, we measure how informative an uncertainty measure is with respect to how well it can classify examples that come from an altogether different dataset as anomalous. However, ideally, we would like this measure of uncertainty to also correspond to the network's own accuracy. To test this, we use the TaxiNet domain as a test case, where we construct a dataset which includes images from all day, morning and afternoon, on a clear day. The training dataset consists only of images from the morning, so this all-day dataset shares some support with the training dataset. Moreover, the differences in these inputs are quite subtle, as the images remain brightly lit, though some shadows appeare in the afternoon. We choose an error threshold, and consider inputs for which the network has an error (measured in Mahalanobis distance) less than this threshold to be ``in-distribution'' and those where the network has a high error to be ``out-of-distribution.'' 

Figure \ref{fig:error-based} shows the results of this experiment. We see that even in this setup, \algName{} produces very high AUROC scores, similar to Deep Ensembles, and outperforms all post-training baselines. In this experiment, we find that applying \algName{} to a single pre-trained DNN almost perfectly characterizes the DNNs domain of competency.

%% file: results_table.tex
\begin{table*}[t]
    \centering
    \footnotesize
    \begin{tabular}{llr@{\hspace{0pt}}lr@{\hspace{0pt}}lr@{\hspace{0pt}}l}
    \toprule
    \textbf{Domain} & \textbf{Method} &
    \multicolumn{2}{c}{\textbf{Runtime} (ms)} &
    \multicolumn{2}{c}{\textbf{AUROC}} &
    \multicolumn{2}{c}{\textbf{AUPR}} \\
    \midrule 
    \multirow{6}{*}{Wine (regression)} & \textbf{\algName{}}& $1.214$ & $\pm0.009$& $0.966$ & $\pm0.002$& $0.962$ & $\pm0.003$ \\
& \textbf{\algName{} (LL)} & $1.023$ & $\pm0.007$& $0.963$ & $\pm0.002$& $0.961$ & $\pm0.002$ \\
& Deep Ensemble & $1.985$ & $\pm0.008$& $0.870$ & $\pm0.004$& $0.896$ & $\pm0.003$ \\
& Local Ensemble & $0.983$ & $\pm0.007$& $0.933$ & $\pm0.002$& $0.931$ & $\pm0.002$ \\
& KFAC Laplace & $6.226$ & $\pm0.043$& $0.945$ & $\pm0.002$& $0.949$ & $\pm0.002$ \\
& Naive & $0.244$ & $\pm0.002$& $0.500$ & $\pm0.000$& $0.750$ & $\pm0.000$ \\
\midrule
    \multirow{6}{*}{Rotated MNIST (regression)} & \textbf{\algName{}}& $1.906$ & $\pm0.011$& $0.909$ & $\pm0.004$& $0.920$ & $\pm0.003$ \\
& \textbf{\algName{} (LL)} & $1.170$ & $\pm0.026$& $0.952$ & $\pm0.002$& $0.940$ & $\pm0.004$ \\
& Deep Ensemble & $3.794$ & $\pm0.012$& $0.788$ & $\pm0.004$& $0.789$ & $\pm0.004$ \\
& Local Ensemble & $1.726$ & $\pm0.010$& $0.677$ & $\pm0.005$& $0.674$ & $\pm0.005$ \\
& KFAC Laplace & $11.650$ & $\pm0.016$& $0.779$ & $\pm0.004$& $0.775$ & $\pm0.005$ \\
& Naive & $0.477$ & $\pm0.004$& $0.500$ & $\pm0.000$& $0.750$ & $\pm0.000$ \\
    \midrule
    \multirow{6}{*}{TaxiNet (regression)} & \textbf{\algName{}} & $18.074$ & $\pm0.018$& $0.995$ & $\pm0.000$& $0.994$ & $\pm0.001$ \\
& \textbf{\algName{} (LL)} & $7.344$ & $\pm0.022$& $1.000$ & $\pm0.000$& $1.000$ & $\pm0.000$ \\
& Deep Ensemble & $30.786$ & $\pm0.026$& $1.000$ & $\pm0.000$& $1.000$ & $\pm0.000$ \\
& Local Ensemble & $13.857$ & $\pm0.024$& $0.842$ & $\pm0.003$& $0.820$ & $\pm0.004$ \\
& KFAC Laplace & $106.839$ & $\pm5.409$& $0.994$ & $\pm0.000$& $0.994$ & $\pm0.000$ \\
& Naive & $2.134$ & $\pm0.007$& $0.500$ & $\pm0.000$& $0.750$ & $\pm0.000$ \\
    \midrule
    \multirow{6}{*}{Binary MNIST (classification / logistic)} & \textbf{\algName{}}& $1.679$ & $\pm0.007$& $0.983$ & $\pm0.001$& $0.982$ & $\pm0.001$ \\
& \textbf{\algName{} (LL)}& $1.185$ & $\pm0.008$& $0.985$ & $\pm0.001$& $0.985$ & $\pm0.001$ \\
& Deep Ensemble & $3.258$ & $\pm0.011$& $0.982$ & $\pm0.001$& $0.981$ & $\pm0.001$ \\
& Local Ensemble & $3.111$ & $\pm0.007$& $0.974$ & $\pm0.001$& $0.971$ & $\pm0.001$ \\
& KFAC Laplace & $9.566$ & $\pm0.046$& $0.972$ & $\pm0.001$& $0.970$ & $\pm0.001$ \\
& Naive & $0.492$ & $\pm0.002$& $0.972$ & $\pm0.001$& $0.970$ & $\pm0.001$ \\
    \midrule
    \multirow{6}{*}{MNIST (classification / softmax} & \textbf{\algName{}} & $6.093$ & $\pm0.005$& $0.963$ & $\pm0.001$& $0.961$ & $\pm0.001$ \\
& \textbf{\algName{} (LL)}& $4.309$ & $\pm0.008$& $0.964$ & $\pm0.001$& $0.966$ & $\pm0.002$ \\
& Deep Ensemble& $3.179$ & $\pm0.013$& $0.965$ & $\pm0.001$& $0.969$ & $\pm0.001$ \\
& Local Ensemble & $6.258$ & $\pm0.014$& $0.957$ & $\pm0.001$& $0.956$ & $\pm0.002$ \\
& KFAC Laplace & $9.631$ & $\pm0.012$& $0.957$ & $\pm0.002$& $0.961$ & $\pm0.002$ \\
& Naive & $0.436$ & $\pm0.003$& $0.958$ & $\pm0.002$& $0.962$ & $\pm0.002$ \\
    \midrule
    \multirow{5}{*}{CIFAR10 (classification / softmax)} & \textbf{\algName{}}& $140.624$ & $\pm0.330$& $0.908$ & $\pm0.002$& $0.864$ & $\pm0.004$ \\
& \textbf{\algName{} (LL) }& $56.677$ & $\pm0.177$& $0.932$ & $\pm0.002$& $0.905$ & $\pm0.003$ \\
& Local Ensembles & $152.154$ & $\pm0.179$& $0.919$ & $\pm0.002$& $0.881$ & $\pm0.003$ \\
& KFAC Laplace & $840.195$ & $\pm1.364$& $0.920$ & $\pm0.002$& $0.915$ & $\pm0.002$ \\
& Naive & $10.929$ & $\pm0.017$& $0.922$ & $\pm0.002$& $0.917$ & $\pm0.003$ \\
    \bottomrule
    \end{tabular}
    \caption{Full Numerical Results. For each domain, we apply each method to output an uncertainty score on both in and out-of-distribution inputs. We evaluate the performance in terms of runtime per example, and the area under the ROC curve (AUROC) and the area under the precision-recall curve (AUPR). For the latter two metrics, 95\% confidence bounds are produced by bootstrapping.}
    \label{tab:my_label}
\end{table*}